\documentclass[conference]{IEEEtran}
\IEEEoverridecommandlockouts
\usepackage{cite}
\usepackage{amsmath,amssymb,amsfonts}
\usepackage{algorithmic}
\usepackage{graphicx}
\usepackage{textcomp}
\usepackage{xcolor}
\usepackage[utf8]{inputenc} 
\usepackage[T1]{fontenc}    
\def\BibTeX{{\rm B\kern-.05em{\sc i\kern-.025em b}\kern-.08em
    T\kern-.1667em\lower.7ex\hbox{E}\kern-.125emX}}
\usepackage{natbib}

\newcommand{\SpeckleNN}{SpeckleNN}

\newcommand{\skopi}{skopi}

\begin{document}

\title{\SpeckleNN{}: A unified embedding for real-time speckle pattern
classification in X-ray single-particle imaging with limited labeled
examples}

\author{\IEEEauthorblockN{Cong Wang}
\IEEEauthorblockA{\textit{Linac Coherent Light Source} \\
\textit{SLAC National Accelerator Laboratory}\\
cwang31@slac.stanford.edu}
\and
\IEEEauthorblockN{Eric Florin*\thanks{* Performed this work in
a previous role at SLAC National Accelerator Laboratory}}
\IEEEauthorblockA{\textit{Computer Science and Engineering} \\
\textit{UCLA}\\
}
\and
\IEEEauthorblockN{Hsing-Yin Chang}
\IEEEauthorblockA{\textit{Linac Coherent Light Source} \\
\textit{SLAC National Accelerator Laboratory}\\
iris@slac.stanford.edu}
\and
\IEEEauthorblockN{Jana Thayer}
\IEEEauthorblockA{\textit{Linac Coherent Light Source} \\
\textit{SLAC National Accelerator Laboratory}\\
jana@slac.stanford.edu}
\and
\IEEEauthorblockN{Chun Hong Yoon}
\IEEEauthorblockA{\textit{Linac Coherent Light Source} \\
\textit{SLAC National Accelerator Laboratory}\\
yoon82@slac.stanford.edu}
}

\maketitle

\begin{abstract}
With X-ray free-electron lasers (XFELs), it is possible to determine the
three-dimensional structure of noncrystalline nanoscale particles using X-ray
single-particle imaging (SPI) techniques at room temperature.  Classifying SPI
scattering patterns, or ``speckles", to extract single hits that are needed for
real-time vetoing and three-dimensional reconstruction poses a challenge for
high data rate facilities like European XFEL and LCLS-II-HE.  Here, we introduce
\SpeckleNN{}, a unified embedding model for real-time speckle pattern
classification with limited labeled examples that can scale linearly with
dataset size.  Trained with twin neural networks, \SpeckleNN{} maps speckle
patterns to a unified embedding vector space, where similarity is measured by
Euclidean distance.  We highlight its few-shot classification capability on new
never-seen samples and its robust performance despite only tens of labels per
classification category even in the presence of substantial missing detector
areas.  Without the need for excessive manual labeling or even a full detector
image, our classification method offers a great solution for real-time
high-throughput SPI experiments.
\end{abstract}


\section{Introduction}

Single-particle imaging (SPI) with X-ray free-electron lasers (XFELs) is a
promising method for determining the three-dimensional structure of
noncrystalline nanoscale particles at room temperature.  In SPI experiments,
femtosecond coherent X-ray beams strike biomolecules injected into the beam path,
causing radiation damage-free scattering of the samples inflicted by the intense
X-rays.  This way of collecting scattering datasets is known as
\textit{diffraction before destruction}
\citep{neutzePotentialBiomolecularImaging2000,
chapmanFemtosecondDiffractiveImaging2006,seibertSingleMimivirusParticles2011,
aquilaLinacCoherentLight2015,reddyCoherentSoftXray2017a}.  Such scattering
patterns are also referred to as ``speckles" due to its grainy appearance.  A
single particle of interest can then be reconstructed by algorithms, such as EMC
\citep{lohReconstructionAlgorithmSingleparticle2009,
ayyerDragonflyImplementationExpand2016} and M-TIP
\citep{donatelliReconstructionLimitedSingleparticle2017,
changScalingAccelerationThreedimensional2021}, from hundreds to tens of
thousands of speckle patterns.

In today's SPI experiments, speckle patterns form in four main categories,
depending on what interacts with the X-ray pulse at the point of interaction.  A
large fraction of X-ray pulses may miss the target particle, e.g.
\citep{shiEvaluationPerformanceClassification2019} reported a 98\% of the pulses
did not interact with the sample, resulting in no scattering pattern defined as
a no-hit.  In contrast, a speckle pattern is labeled as a single-hit when X-ray
photons collide with one and only one sample particle. Similarly, a multi-hit
happens when an X-ray pulse intersects with two or more sample particles. In
some cases, X-ray pulses might also hit objects that are not the sample of
interest in the delivery medium and those speckle patterns are defined as
non-sample-hit.  The main goal of this work is to provide an efficient solution
to identify single-hit speckle patterns in near real-time during data
collection.

Real-time speckle pattern classification for SPI experiments is a challenge
faced by high data rate facilities like European XFEL and LCLS-II-HE, due to
their need for (1) real-time vetoing to better utilize data storage, and (2) to
enable near real-time feedback of reconstructed electron densities.
Classification algorithm need to scale linearly to handle the vast amount of
data they generate in real-time.  Some pioneering works addressing the challenge
employed unsupervised learning techniques
\citep{yoonUnsupervisedClassificationSingleparticle2011,
giannakisSymmetriesImageFormation2012,schwanderSymmetriesImageFormation2012,
yoonNovelAlgorithmsCoherent2012,
andreassonAutomatedIdentificationClassification2014,
bobkovSortingAlgorithmsSingleparticle2015}.  Those solutions can reveal
underlying clusters of data categories and runs without human labeling, but
require post-human interpretation to achieve reasonable classification results,
such as specifying the decision boundary of single-hit speckle patterns in some
vector space. Also, these algorithms do not scale linearly with the number of
speckle patterns needed for real-time classification.  On the other hand,
supervised learning solutions based on artificial neural network models
\citep{shiEvaluationPerformanceClassification2019,
ignatenkoClassificationDiffractionPatterns2021} scale linearly, but requires
hundreds of labeled examples of the data being collected during beam time and
require additional time for model training which precludes real-time
classification.  Those models are made of largely two components: (1) Some
convolutional neural networks (CNN) for spatial feature extraction; (2) Some
fully connected networks (FCN) for compressing CNN features into probability
distribution of possible outcomes.  These models demonstrated good performance
on speckle patterns of one single-particle sample, bacteriophage PR772, which is
an important step towards the goal of near real-time particle classification.
But when it comes to a different sample, they will have to be retrained on
hundreds of labeled speckle patterns.  Notably, it is not a lack of computing
power that precludes these model from working in near real-time, since we can
run deep learning models on supercomputers with modern graphics processing units
(GPUs).  The bottleneck is speckle pattern labeling.  It is by no means a
trivial task to label speckle patterns, especially at scale, even for experts in
the field.  Therefore, we need solutions that enable neural network models to
effectively classify speckle patterns without excessive manual labeling.  

We aim to address the problem of near real-time speckle pattern classification
by converting it into the task of measuring speckle pattern similarities.  To
accomplish this goal, we propose to train a neural network model to learn an
embedding function, capable of mapping speckle patterns into a unified embedding
vector space.  An important property of this vector space is that similarities
can be evaluated by computing the Euclidean distance between any two points in
the space.  Then, we classify unknown examples by comparing them to a few
labeled examples per class in the embedding vector space and assigning the label
of the closest class.

A contrastive approach, known as twin neural networks, is used for training the
unified embedding model.  The main idea is that two identical networks will
extract features from a pair of examples, either with the same label or
different labels.  For two examples with the same label, their Euclidean
distance should be small and vice versa.  Contrastive approaches based on twin
neural networks have achieved early success in computer vision tasks such as
signature verification \citep{bromleySignatureVerificationUsing1993} and face
verification \citep{chopraLearningSimilarityMetric2005,
schroffFaceNetUnifiedEmbedding2015}.  Moreover, two twin neural networks can
work together to collectively train the underlying embedding network by
minimizing the Euclidean distance between identically labeled examples, while
simultaneously maximizing the Euclidean distance between differently labeled
examples.  This approach is also referred to as triplet networks detailed in the
face verification model \citep{schroffFaceNetUnifiedEmbedding2015}.

In this work, we present \SpeckleNN{} as a unified embedding network for
classifying speckle patterns in real-time X-ray single particle imaging.
Concretely, two classification solutions are proposed, one with offline training
and one with online training:

\begin{itemize}

    \item First, we show that \SpeckleNN{} accurately classifies speckle
    patterns with few labeled examples per category (e.g. 5) by learning a
    unified embedding function from a vast number of distinct samples (e.g.
    different proteins).  The model can be trained entirely offline or prior to
    data collection, and its classification capability generalizes to new
    samples.

    \item Second, we demonstrate that \SpeckleNN{} achieves accurate and robust
    speckle pattern classification in the presence of missing detector area
    (e.g.  25\% of a pnCCD detector).  The model is designed to be trained
    online or during data collection on a relatively small number of labeled
    examples per category (around 60).  Unlike the offline solution, the online
    training utilized only the speckle patterns from the sample of interest as
    opposed to generalize embedding of multiple samples.

\end{itemize}


\section{Related work}

\subsection{Speckle pattern classification}

The task of single-particle speckle pattern classification often requires expert
knowledge and immense manual effort.  Human-engineered feature extractor and
unsupervised learning came along to tackle this challenge.  For instance,
spectral clustering \citep{yoonUnsupervisedClassificationSingleparticle2011},
principal component analysis (PCA) and support vector machines (SVM)
\citep{bobkovSortingAlgorithmsSingleparticle2015} were used for single-hit
classification.  Geometric machine learning is a supervised learning solution
based on the diffusion map framework that can output a score for how likely a
speckle pattern is a single-hit \citep{cruz-chuSelectingXFELSingleparticle2021}.
More recently, artificial neural network models have become a new avenue for
exploring classification solutions with the advent of capable infrastructures
(GPUs, machine learning frameworks) for model training.
\citep{shiEvaluationPerformanceClassification2019} uses a CNN for feature
extraction and couples its last layer with two additional fully connected (FC)
layers that perform binary classification, which achieved an accuracy of 83.8\%
in predicting single-hits.  More recently, another neural network based hit
classifier is proposed by \citep{ignatenkoClassificationDiffractionPatterns2021}.
They repurposed YOLO (You only look once) deep learning models
\citep{redmonYOLO9000BetterFaster2016, redmonYOLOv3IncrementalImprovement2018}
from detecting objects to classifying speckle patterns.  In fact, these YOLO
models also consist of a CNN spatial feature extractor and several FC layers to
compress features into the probability of classes and location of objects.
However, these models cannot be directly used to classify speckle patterns of
previously unseen single-particle samples without example relabeling and model
retraining.  Their performance with missing detector area is also unknown.  YOLO
models, specifically, come with extra complexities, such as requiring bounding
boxes as labels and increasing computational cost for finding the location of a
speckle pattern.

\subsection{Similarity metrics in a unified embedding vector space}

To the best of our knowledge, there is currently no solution that directly maps
single-particle speckle patterns of distinct samples to a unified vector space,
where similarity is characterized by Euclidean distance.  However, the idea of
introducing similarity metrics in a unified vector space is not new.  One of the
early examples was signature verification using a twin neural network
\citep{bromleySignatureVerificationUsing1993}.  During training, a twin neural
network works on two signatures simultaneously.  During verification, only one
half of the twin neural network is used to map input signatures into a vector
space.  The output embedding will be compared with previously stored signature
embedding in this unified vector space.  The stored embedding that is closer to
the input embedding is considered to share the same label as the input, thereby,
the label of this stored embedding becomes the predicted label of the input.
Similar twin neural networks were later used to train models for face
recognition/verification \citep{chopraLearningSimilarityMetric2005}.  Then,
triplet neural networks \citep{hofferDeepMetricLearning2014} were applied to
further enhance the unified embedding models
\citep{schroffFaceNetUnifiedEmbedding2015} by training essentially two
twin-neural networks on both positive and negative examples simultaneously
instead of only one of them.  In addition to twin neural networks and its
variants, many embedding models have been explored for few-shot classification.
\citep{vinyalsMatchingNetworksOne2017} introduced ``matching networks" that map
queries and supports to a unified vector space with two independent embedding
functions.  \citep{snellPrototypicalNetworksFewshot2017} proposed that a unified
embedding model can be trained with ``prototypical networks" so that the
embedding of unseen inputs is more likely to be closer to the correct
``prototype", defined as the mean of the embedded supports of the same category.


\section{Methods}

\begin{figure}[htbp] 
\includegraphics[width=0.5\textwidth,keepaspectratio]{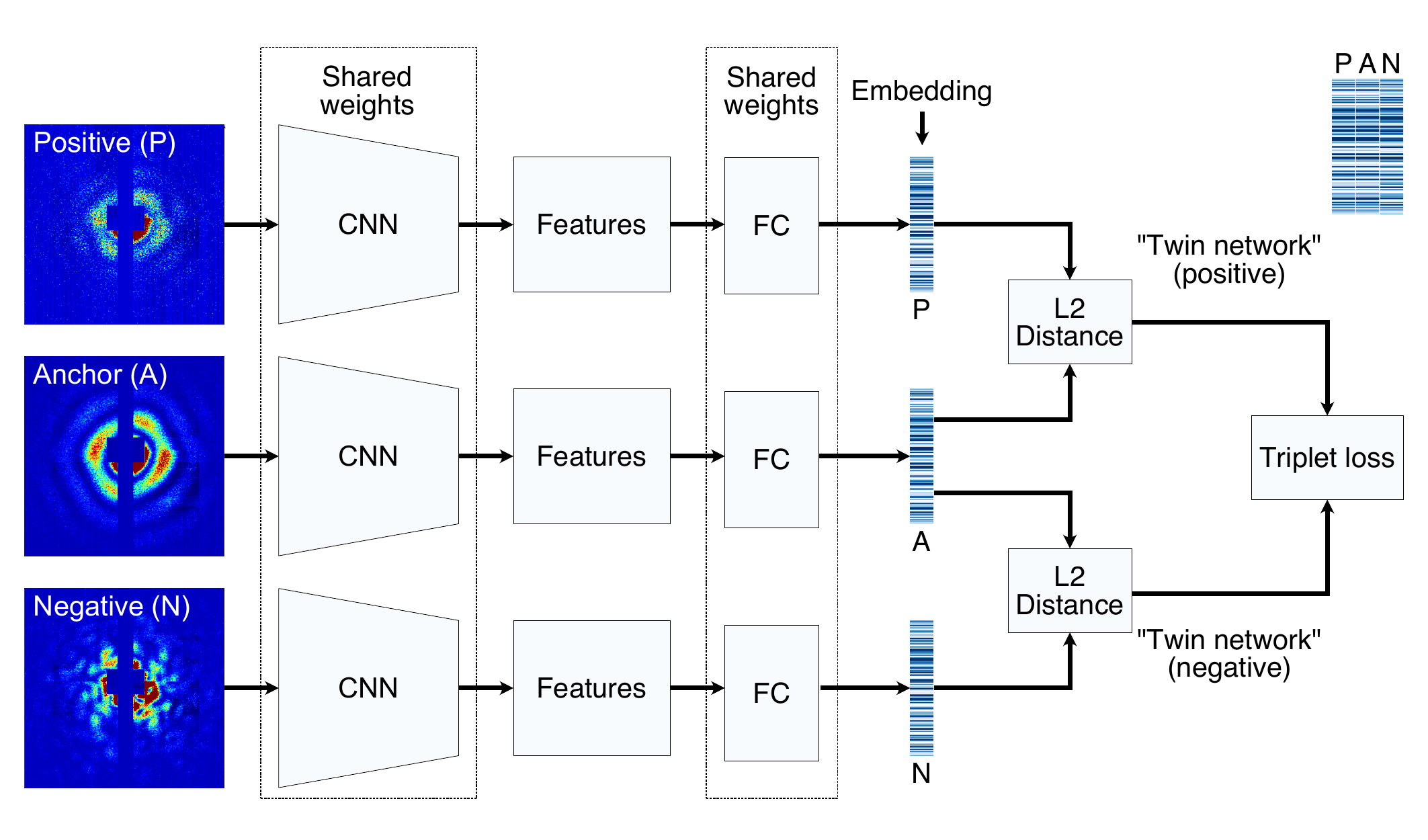}
\caption{ The triplet network architecture for model training.
Three input examples (anchor, positive and negative) are propagated through the
triplet neural network simultaneously.  Anchor and positive share the same label,
thus forming a matching pair.  In contrast, anchor and negative do not share the
same label, thus forming an opposing pair.  The three CNNs and FC layers share
the same weights in the triplet network.  After examples are embedded to a low
dimensional vector space, a triplet loss function is used to simultaneously
maximize similarities between matching embeddings and minimize those between
opposing embeddings.  A side-by-side comparison of three embeddings in a triplet
are annotated at the upper right corner. }
\label{network architecture}
\end{figure}

Our speckle pattern classifier uses a unified embedding model to measure pattern
similarity through Euclidean distance in the embedding space.  This model is
trained by two twin neural networks simultaneously.  One twin neural network
processes two matching examples that share the same label, while the other works
on two opposing examples with different labels.  Such dual twin neural networks
can be further simplified to triplet networks when the matching pair and the
opposing pair share a common example.  The common example is referred to as
anchor, and the matching example and the opposing example are referred to as
positive and negative, respectively.  The complete triplet network architecture
is summarized in Fig.  \ref{network architecture}.  In this section, we present
the details in model training with triplet networks, including the embedding
model, the loss function and the selection of triplet examples.  Additionally,
we outline the steps for speckle pattern classification.

\subsection{The embedding model (the vision backbone)}

\begin{figure}[htbp]
\includegraphics[width=0.5\textwidth,keepaspectratio]
{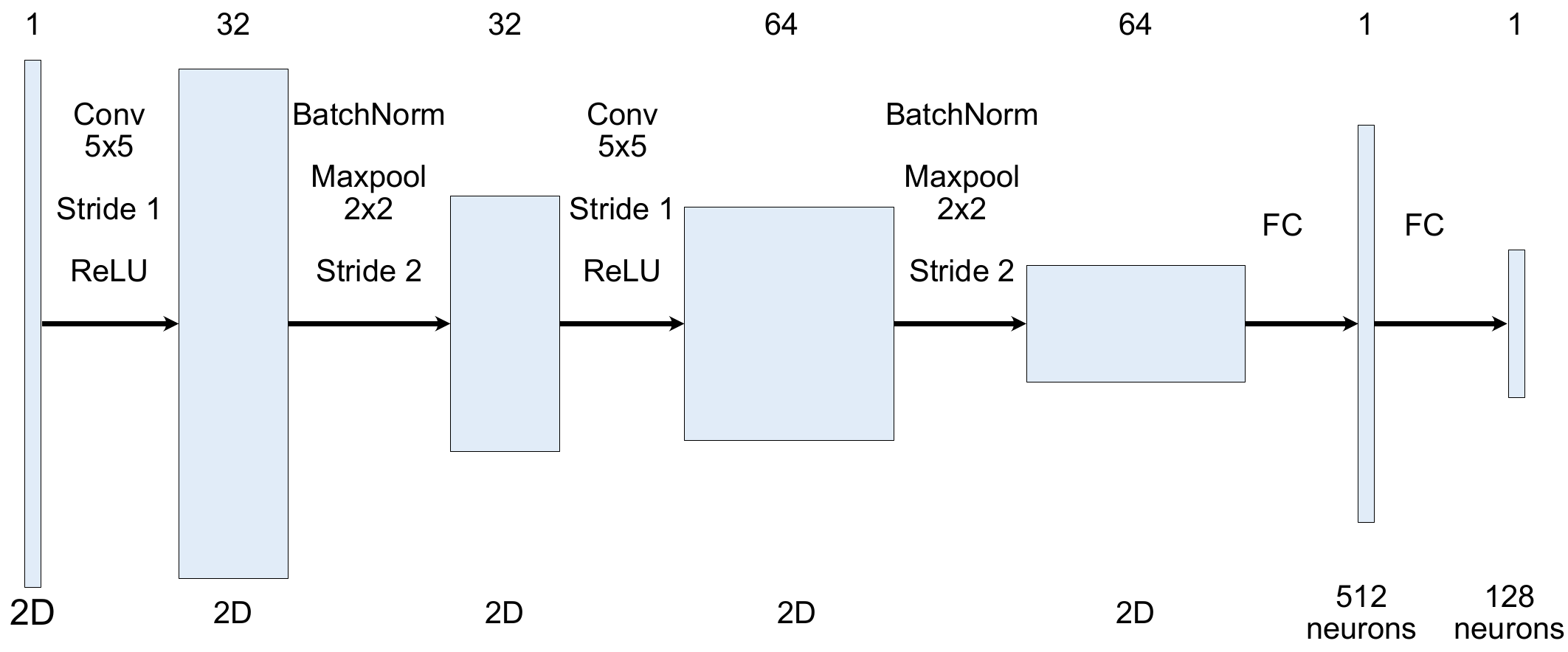}
\caption{The network architecture of the embedding model.  Each shaded rectangle
is a volumetric data representation in the neural network pipeline.  The channel
number of each data representation is marked on the top row.  The type of
spatial dimension, such as 2D tensors or 1D tensor of neurons, is also annotated for
each data representation at the bottom row.  Notably, the initial spatial
dimension of the input may change if cropping and resizing are applied.}
\label{vision backbone architecture}
\end{figure}

Our embedding model consists of two convolutional layers that extract spatial
features and two fully connected layers that compress these features into a low
dimensional vector, or embedding.  The detailed architecture of the embedding
model is delineated in Fig. \ref{vision backbone architecture}.  The first
convolutional layer uses a $5 \times 5$ single channel filter with a stride of
one and no padding.  The second convolutional layer employs a 32-channel $5
\times 5$ filter with a stride of one and no padding.  A ReLU (rectified linear
activation unit) activation function is applied to the outcome of each
convolutional layer, which is followed by a batch normalization layer and a
max-pooling operation performed by a $2\times 2$ filter with a stride of two.
Then, two fully connected layers are used to generate the final embedding with
the size of 128.  This way, speckle patterns are encoded into embeddings in a
low dimensional vector space, where the similarity of two embeddings can be
evaluated by their squared $L2$ distance.

\subsection{Triplet loss function}

We train the embedding model in each twin neural network by using a triplet loss
function described in \citep{schroffFaceNetUnifiedEmbedding2015}.  Each input
consists of a triplet of training examples, which are called anchor $x^a$,
positive $x^p$ and negative $x^n$. Anchor has the same label as positive but not
negative.  Together, $(x^a, x^p)$ forms a matching pair, while $(x^a, x^n)$
forms an opposing pair.  During training, three embedding models (CNN+FC) $f$
with shared weights map each element in a triplet $(x^a, x^p, x^n)$ into a
unified embedding vector space, respectively.  The objective of training is to
separate the two embeddings in each opposing pair by at least a margin of
$\alpha$ from the embeddings in the corresponding matching pair in the vector
space.  Given $N$ triplets, the training objective can be stated as 

\newcommand{\isep}{\mathrel{{.}\,{.}}\nobreak}
\begin{equation} \label{eq:objective}
    \begin{array}{l}
       \|f(x_i^a) - f(x_i^p)\|_2^2 + \alpha < \|f(x_i^a) - f(x_i^n)\|_2^2, \, i = 1 \isep N
    \end{array}
\end{equation}

Meanwhile, we enforce that every embedding has a unit length of one in a
$d$-dimensional vector space, namely $f(x) \in \mathbb{R}^d$ and $\|f(x)
\|_2=1$. It means that any speckle pattern will be mapped to a single point on a
$d$-dimensional hypersphere with the radius of one.  The largest value of a
possible $\alpha$ is 4 in the squared $L2$ norm sense.

To facilitate the training, the objective in Eq. (\ref{eq:objective}) is turned
into the triplet loss function in Eq. (\ref{eq:triplet_loss}).

\begin{equation} \label{eq:triplet_loss}
    \sum_{i=1}^{N} \left[ \alpha + \|f(x_i^a) - f(x_i^p)\|_2^2 - \|f(x_i^a) -
    f(x_i^n)\|_2^2 \right]_+
\end{equation}

where $[\cdot]_+$ returns zero unless the input value is positive.

\subsection{Selection of semi-hard triplets}

\begin{figure}[htbp]
\includegraphics[width=0.5\textwidth,keepaspectratio]
{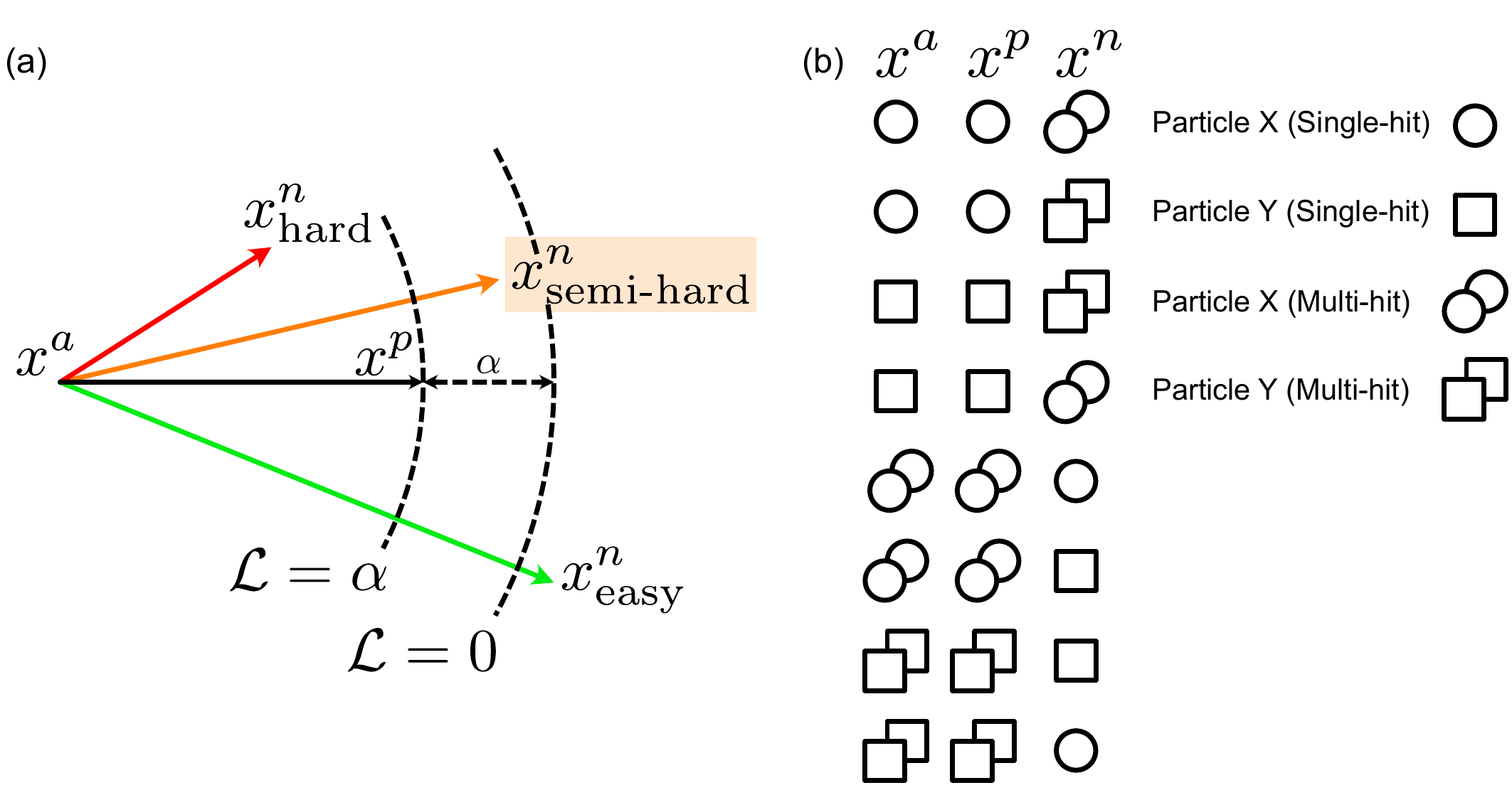} 
\caption{An illustration of the three types of negative examples.  (a) $x^a$
represents an anchor example and $x^p$ is a positive example.  Two arcs in
dashed lines, both centered at $x^a$, are used to divide the embedding space
into three areas.  The inner arc has a radius of $\|f(x_i^a) - f(x_i^p)\|_2^2$,
whereas the outer arc has a radius that is larger by a margin of $\alpha$.
Negative examples will possess three difficulty levels in model training based
on the area where they are situated.  It is considered hard negative example if
it is located within the inner arc, where $\|f(x_i^a) - f(x_i^n)\|_2^2 <
\|f(x_i^a) - f(x_i^p) \|_2^2$.  On the contrary, it is considered an easy
negative example when it goes outside the outer arc, where $\|f(x_i^a) - f(x_i^n)
\|_2^2 - \|f(x_i^a) - f(x_i^p) \|_2^2 > \alpha$.  Lastly, it becomes a semi-hard
negative example when it resides in the area bound between the two arcs.
Moreover, the loss function results in $\mathcal{L} = \alpha$ and $\mathcal{L} =
0$ when $x^n_{\text{semi-hard}}$ is on the inner arc and outer arc,
respectively.  Our model training will pull $x^n_{\text{semi-hard}}$ close to
the outer arc as much as possible, namely minimizing the loss.  (b) An
illustration of possible semi-hard scenarios when two unique single-particle
samples are involved.}
\label{fig: semi-hard} 
\end{figure}

A triplet can be randomly selected in three steps: (1) Randomly choose a class;
(2) Randomly sample two unique examples from the chosen class;  (3) Randomly
sample one example from any class other than the chosen class.  This method,
despite being easy to implement, might not deliver fast convergence when there
are too many easy triplets.  To explain it in detail, we consider three kinds of
triplets that might exist during model training: easy, semi-hard and hard, as
shown in Fig.  \ref{fig: semi-hard}(a).  In an easy triplet, the negative
example is already separated by at least a margin of $\alpha$ than the positive
example.  In a hard triplet, the negative example is actually closer to the
anchor than the positive example.  In a semi-hard triplet, the negative example
is farther away from the anchor than the positive example with a margin smaller
than $\alpha$.  The problem with easy triplets is that they contribute to zero
in the triplet loss, and thus the model weights will be adjusted only according
to other triplets, namely semi-hard and hard triplets.  The problem with too
many hard examples is that they mostly constitute only a small fraction of the
whole population.  If the optimization prioritizes separating them from their
corresponding anchors, the loss function will more likely get stuck in some bad
local minima.  Therefore, selecting semi-hard triplets for training is
important.  This strategy does not mean to ignore hard examples at all.  Instead,
once a hard example is pulled into the semi-hard zone, optimization can further
drive them into the easy zone. This allows majority of the negative examples to
stay away from their anchor by a considerable margin $\alpha$.  From a practical
standpoint, the selection of semi-hard triplets is done at the mini-batch level,
where our model randomly selects a triplet that satisfies the following
condition.  

\begin{equation}\label{eq: semi-hard condition}
    \begin{aligned}[b]
    &\|f(x_i^a) - f(x_i^p)\|_2^2 \;<\; \|f(x_i^a) - f(x_i^n)\|_2^2 \;<\;
    \|f(x_i^a) - f(x_i^p)\|_2^2 + \alpha, \\
    &i = 1 \isep N
    \end{aligned}
\end{equation}

However, complexity arises with semi-hard selections involving multiple
single-particle samples.  We choose to select random anchor $x^a$ and positive
$x^p$ from the same sample with the same label, with the negative $x^n$ selected
from any sample with a different label.  Under this selection scheme, we lay out
all scenarios for semi-hard selections when two unique single-particle samples
(Particle X and Particle Y) are present, as shown in Fig.  \ref{fig: semi-hard}(b).  

\subsection{Optimization}

We trained our neural network models using Adam
\citep{kingmaAdamMethodStochastic2017} with a learning rate of $10^{-3}$.  The
model weights are initialized to random values from a Gaussian probability
distribution with a mean of $0.0$ and a standard deviation of $0.2$.

\subsection{Data augmentation}

Data augmentation is widely used in many machine learning tasks to address
limitations imposed by expensive human-labeling and improve model performance.
In essence, ``a data-augmentation is worth a thousand samples"
\citep{balestrieroDataAugmentationWorthThousand2022}.  We applied four data
augmentation strategies to each speckle pattern in our dataset, including random
in-plane rotation, random masking, random zooming and random shifting in both
horizontal and vertical directions.  Random in-plane rotation mimics the effect
of single-particle rotation.  Random masking covers some area of a speckle
pattern with constant-value pixel intensities to be more robust to bad pixels
and parasitic scattering.  Random zooming and random shifting enforce the model
to learn features independent of detector distance, X-ray wavelength, and X-ray
beam center.  These data augmentation strategies expand the data distribution
for the model without manual labeling.  

An important caveat when applying data augmentation is to partition the data
into training set and test set before the augmentation.  Otherwise, it will
lead to ``data leakage'' as explained in
\citep{kapoorLeakageReproducibilityCrisis2022}.  One consequence of ``data
leakage'' is the deceptively good model predictive performance measured on a
test set that already contains data augmented or ``leaked'' from the training
set.  In other words, the ``good'' performance will be mostly caused by model
memorization or overfitting rather than generalization.

\subsection{Four steps in classification}

\begin{figure}[htbp]
\includegraphics[width=0.5\textwidth,keepaspectratio]
{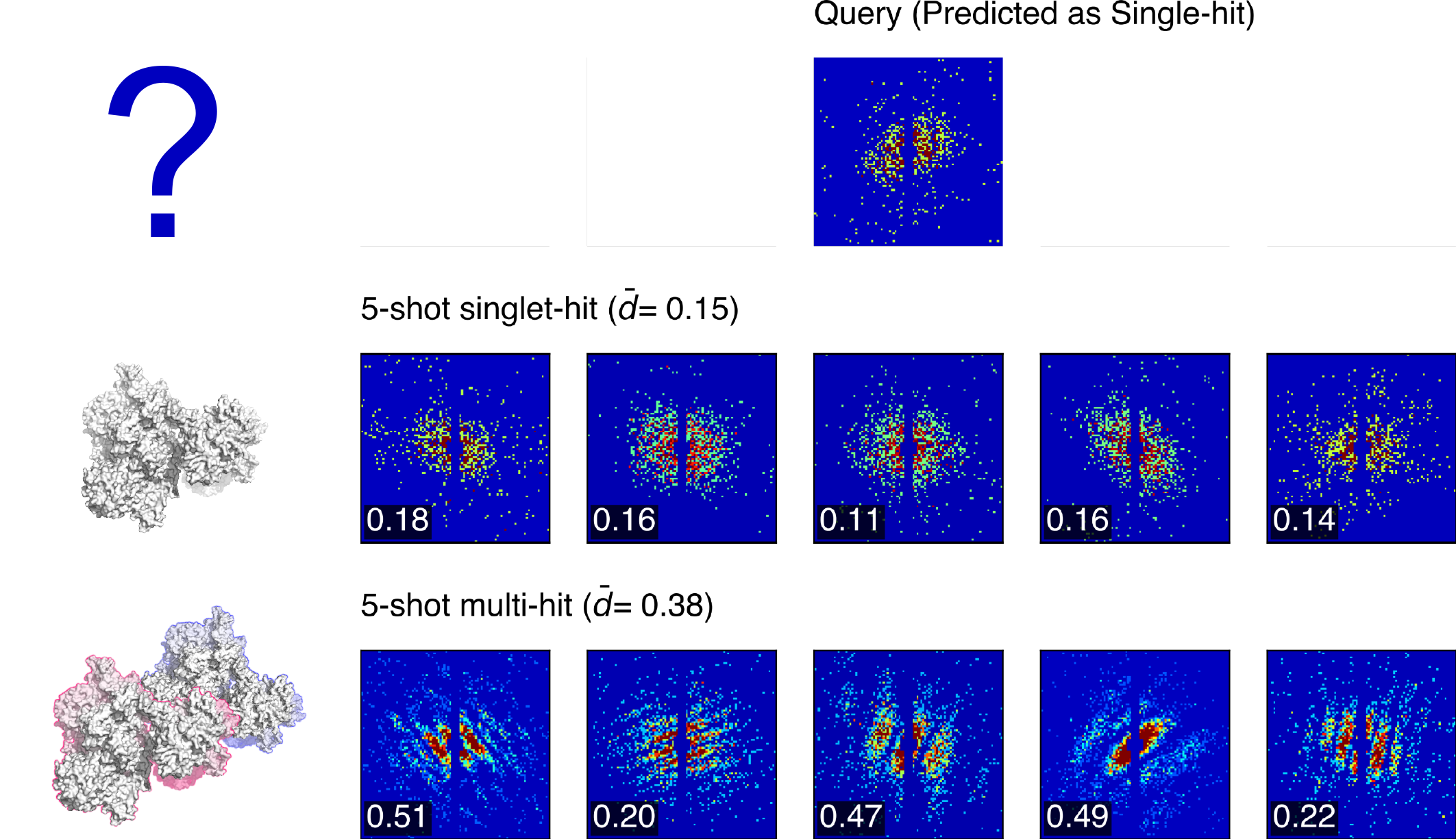}
\caption{An illustration of a 2-way 5-shot classification.  The speckle patterns
in this figure are simulated from a type VI secretion system (PDB entry: 6N38).
A queried speckle pattern is shown in the first row that has a ground truth
label of single-hit.  The second and third rows represent the single-hit and
multi-hit support sets, respectively.  5 patterns are used in each support set.
The query-to-support distance is annotated at the bottom left corner of each
speckle pattern.  The average query-to-support distance is denoted as $\bar{d}$.
The single-hit class has a shorter $\bar{d}$ than the multi-hit class,
resulting in a single-hit label for the query pattern.  Additionally, a question
mark in the first row indicates a to-be-determined label of a queried particle.
To illustrate a single-hit example, we use a simple cartoon representation of a
6N38 molecule in the second row.  Likewise, a multi-hit example is depicted by a
cartoon representation of two 6N38 molecules with distinctly colored outlines.}
\label{fig : 5-shot demo}
\end{figure}

Our model maps speckle patterns into a unified embedding space, without directly
predicting labels.  Instead, label prediction is performed in a
query-against-support manner, that is, comparing inputs (queries) to labeled
examples (supports).  This approach is also referred to few-shot classification,
often implying novel classes for queries and supports.  In an $N$-way $X$-shot
classification, $N$ is the number of classes and $X$ is the number of labeled
examples per class.  The classification takes four steps: (1) We embed an
unknown input speckle pattern and all support examples in a unified embedding
space.  (2) We calculate the Euclidean distances from the input to every support
example.  (3) We average all distances by class.  (4) We rank all classes by
average distance and select the class with the shortest distance as the label of
the unknown speckle pattern.  Fig. \ref{fig : 5-shot demo} demonstrates an
example of 2-way 5-shot classification.

\section{Experiments}

The ultimate goal of \SpeckleNN{} is to accurately classify speckle patterns.
Our unified embedding model facilitates the conversion of the speckle pattern
classification problem into a set of similarity measures.  Here we demonstrate
two classification solutions, one with offline training and one with online
training.  Offline training requires past experimental data to train the model,
which is then directly applied to classification of speckle patterns in future
experiments.  However, online training trains the model solely on newly
collected data in an ongoing experiment.  Both offline and online training are
important for SPI experiments in high data rate facilities.  Offline training
offers a ready-to-use model for a wide range of samples, while online training
provides a potentially more accurate experiment-specific model for the sample
of interest.  

As mentioned, training offline model requires speckle patterns from a variety of
distinct samples.  To this date, successful 3D reconstructions from single
particle datasets are primarily from large viruses, such as mimivirus
\citep{seibertSingleMimivirusParticles2011}, rice dwarf virus
\citep{janoshajduCoherentDiffractionSingle2016}, and bacteriophage PR772
\citep{liDiffractionDataAerosolized2020}.  Therefore, we elected to demonstrate
the offline-trained model on simulated data.  Meanwhile, our model is designed
to work on one sample of interest when trained online.  We chose to use real
bacteriophage PR772 data collected at LCLS for demonstration of model training
and testing.

\subsection{Offline training}

\subsubsection{Dataset}

\begin{figure}[htbp]
\includegraphics[width=0.5\textwidth,keepaspectratio]
{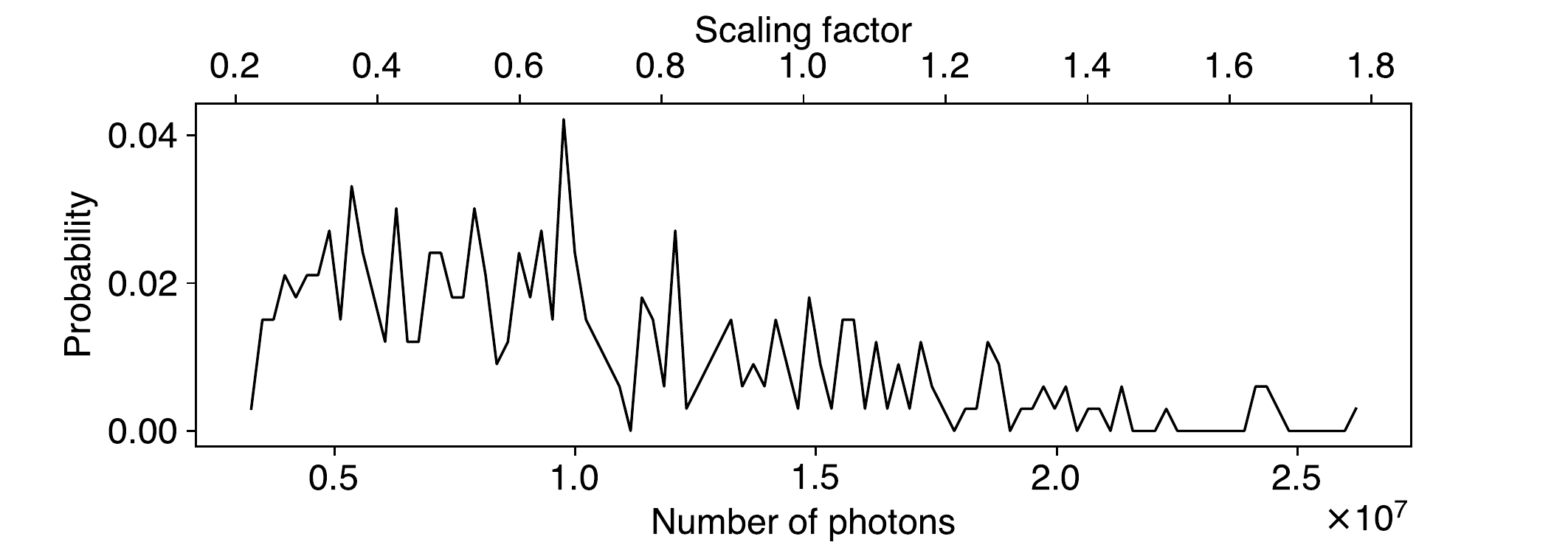}
\caption{ Probability distribution of photon numbers in 332 single-hit speckle
patterns, obtained from LCLS experiment of bacteriophage PR772 at the AMO
instrument (Experiment ID: amo06156, Run numbers: 90, 91, 94, 96 and 102)
\citep{liDiffractionDataAerosolized2020}.  Dividing by mean photon numbers
($\approx 1.5 \times 10^7$) produces scaling factors specified in the upper
x-axis. }
\label{fig : photon dist}
\end{figure}

We randomly selected 100 Protein Data Bank (PDB) entries for model training and
validation, and another 345 PDB entries for model testing.  The number of atoms
in those PDB entries ranges from $10^4$ to $10^5$.  For every PDB entry, we
simulated 400 speckle patterns, with 100 per hit category, from randomly
oriented particles in the form of single-hit, double-hit, triple-hit and
quadruple-hit.  For training purposes, we keep the single-hit label but relabel
the rest as multi-hit.  The beam profile employed in the simulation has a radius
of 0.5 $\mu m$ and a photon energy of 1.66 $keV$, and contains $10^{12} $
photons per pulse.  We simulated all speckle patterns using \skopi{}
\citep{peckSkopiSimulationPackage2022} on a square detector with a dimension of
$172 \times 172$ pixels.  To replicate the conditions similar to real
experiments, we firstly applied a $6 \times 8$ pixel binary mask mimicking a
beam stop at the center and another $172 \times 4$ pixel binary mask resembling
a gap dividing a detector in the middle.  Then, X-ray fluence jitter and shot
noise were introduced to the dataset.  Specifically, during model training, we
introduce fluence jitter by rescaling the intensity of speckle patterns with a
multiplier sampled from an experimental photon number distribution shown in Fig.
\ref{fig : photon dist}.  We also added Gaussian noise with zero mean and 0.15
standard deviation.  Each speckle pattern is also cropped at the center with a
window size of $96 \times 96$.  Data augmentation described in the method
section was applied subsequently.

The speckle patterns simulated from the 100 PDB entries are split into $70\%$ for
training and $30\%$ for validation.  With data augmentation, we obtained $30,000$
speckle patterns for model training and another $10,000$ for model validation.
We found applying data augmentation can add significant latency to the training
process, so decided to cache all speckle patterns into CPU memory.  But it is
possible to pack even more data for model training through better practices,
such as applying data augmentation to a new batch of data while training on the
previous batch is still underway.  

The test set was formed by simulating speckle patterns from 345 PDB entries.
For model prediction, we generated $1,000$ speckle patterns for each PDB entry
through random in-plane rotation as the only data augmentation strategy.  The
main reason is that speckle patterns aren't subject to random masking, random
shifting or random zooming as long as the experimental setup stays unchanged.
We computed a confusion matrix for each PDB entry and reported accuracy and F-1
scores in the following results.

\subsubsection{Performance and photon fluence}

X-ray photon fluence jitter is often present in SPI speckle patterns.  To
illustrate how fluence jitter affects our model performance, we scanned a range
of fluence scaling factors from $10^{-2}$ to $10^2$ by multiplying by $10^{0.5}$
at each step.  These scaling factors are then applied to simulated speckle
patterns in the test set.  Meanwhile, we also measured model performance in
three unique few-shot classification scenarios, including 1-shot, 5-shot and
20-shot.  At the baseline photon fluence, our model achieves an average accuracy
of 84.3\%, 89.1\% and 89.6\% with 1-shot, 5-shot and 20-shot classification,
respectively.  The corresponding F-1 scores are 82.0\%, 87.4\% and 87.9\%,
respectively.  Accuracy and F-1 scores both rise in response to the increase of
photon fluence, and converge at about $10^{0.6} (\approx 4.0) \times$ the
baseline photon fluence.  

There are two main lessons we learned from this result.  Firstly, the
improvement in classification diminishes quickly with increase in support size.
For example, 5-shot classification has a much better performance than 1-shot
classification, but it delivers a comparable performance as 20-shot
classification.  If we were to deploy \SpeckleNN{} at an undergoing experiment,
it would be more appealing to only label 5 examples per category rather than 1
or 20 examples.  Secondly, as free electron laser technology improves in peak
brightness\citep{liFemtosecondTerawattHardXRay2022}, the benefit of having higher
photon fluence can directly improve \SpeckleNN{}'s accuracy.  Interestingly, a
$4 \times$ photon fluence improvement is sufficient to maximize model
performance.

\begin{figure}[htbp]
\includegraphics[width=0.5\textwidth,keepaspectratio]
{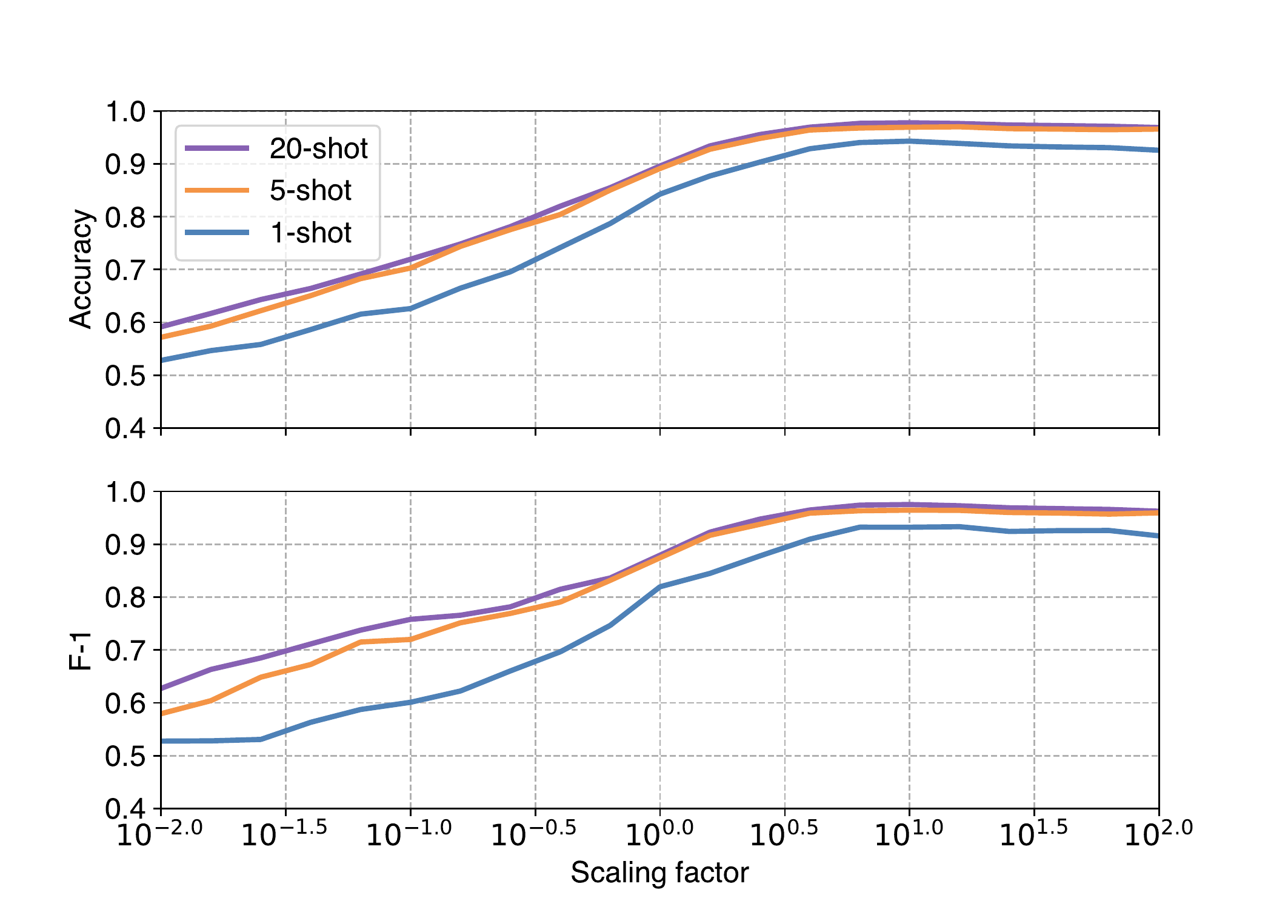}
\caption{ 2-way $X$-shot ($X = 1, 5, 20$) classification performance of our
model, as measured by accuracy and F-1 scores under a range of fluence
conditions.  The baseline fluence is $10^{12}$ photons per X-ray pulse. }
\label{fig : perf vs scale}
\end{figure}

\subsubsection{Performance and particle size}

PDB entries vary significantly in particle size from hundreds to millions of
atoms, and their sizes are unevenly distributed.  It is not a good practice to
form a training dataset by randomly picking PDB entries, which might result in
many entries aggregated within a small size range.  Models trained on such
dataset might perform well only on the highly populated size ranges but less so
otherwise.  We pick roughly equal number of PDB entries from 20 evenly spaced
intervals from $10^4$ to $10^5$ atoms.  The distribution of PDB entries over
particle size used in our model training is visualized in Fig. \ref{fig : perf
vs atom number}(a).  The PDB entries used in the test set were also sampled from
the same size range.

\begin{figure}[htbp]
\includegraphics[width=0.5\textwidth,keepaspectratio]
{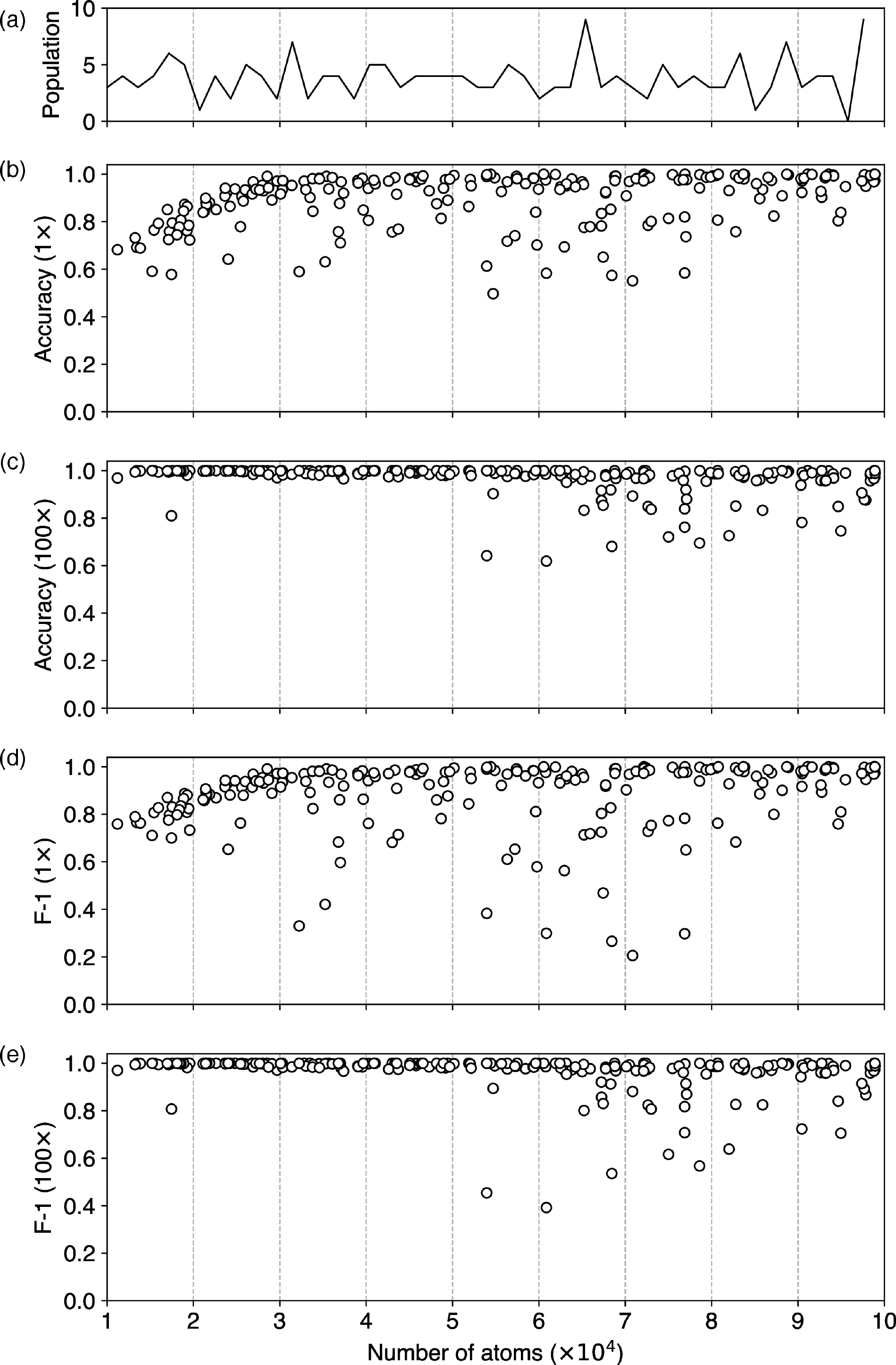}
\caption{ (a) The distribution of particle size, as characterized by the number
of atoms, among PDB entries used in training the model.  (b-e) 2-way 20-shot
classification accuracy (b,c) and F-1 scores (d,e) plotted against number of
atoms in each PDB entry under two respective fluence conditions, 1 $\times$ the
baseline fluence and 100 $\times$ the baseline fluence.  These conditions are
labeled on the y-axes. }
\label{fig : perf vs atom number}
\end{figure}

We pointed out that particle size is a limiting factor in model prediction but
conditioned on photon fluence.  As shown in Fig. \ref{fig : perf vs atom number}
(b), the average test accuracy of our model at the baseline photon fluence
($1\times$ condition) is positively correlated with particle size in the region
of $1 \times 10^4$ to $3 \times 10^4$ atoms.  The average test accuracy becomes
stable when particle size is larger than $3\times 10^4$ despite the presence of
outliers.  Likewise, we repeated model prediction at $100\times$ larger photon
fluence.  No more size dependent correlation is observed across the whole size
range, as shown in Fig. \ref{fig : perf vs atom number}(c).  Similar observation
is also present in F-1 scores as shown in Fig. \ref{fig : perf vs atom number}(d)
and Fig. \ref{fig : perf vs atom number}(e).

\subsection{Online training}

\subsubsection{Dataset}

We obtained speckle data from an LCLS experiment of bacteriophage PR772
collected at the AMO instrument (Experiment ID: amo06156, Run numbers: 90, 91,
94, 96 and 102) \citep{liDiffractionDataAerosolized2020}.  We prepared 332
single-hit, 165 multi-hit and 98 non-sample-hit patterns to form our source
dataset.  Non-sample-hit patterns do not exist in simulated data and are unique
to experimental data, which are largely caused by parasitic scattering.  We
split data into 50\% training set, 25\% validation set and 25\% test set.  All
speckle patterns were subject to data augmentation, specifically random rotation
and random masking, from the source dataset.  It is worth noting that we limit
only 40 and 20 labeled examples per category for model training and validation,
respectively.  The purpose of imposing this restriction is to reproduce the
shortage of labeled examples in a real experiment.  However, we applied data
augmentation to boost the number of examples per category.  Consequently, there
were 1374 non-sample-hit, 1288 single-hit and 1338 multi-hit patterns for model
training, while there were 1359 non-sample-hit, 1352 single-hit and 1289
multi-hit patterns for model validation.

\subsubsection{Robust classification despite missing detector area}

To illustrate robust classification despite missing detector area, we need to
choose a baseline model as a reference point for comparison.  We decided to use
\citep{shiEvaluationPerformanceClassification2019}'s model for this purpose which
we will refer to as Shi19 from here on out.  It consists of a CNN vision
backbone and a multi-layer perceptron (MLP) that outputs probabilities of each
label.  It reportedly achieved 83.8\% accuracy in predicting single-hits.  We
reimplemented Shi19 model in PyTorch for measuring its performance.  It is worth
noting that we need to relabel non-sample-hit and multi-hit as non-single-hit to
accommodate the training of Shi19 model, as it was initially designed for binary
classification.  We still used the original three labels to train our model, and
only relabel them when producing compatible confusion matrices.

Performance comparison between models was conducted for two scenarios: (1) 100\%
detector area is available; (2) 25\% detector area is available.  The second
scenario is more commonly seen in modular detectors, where certain area of the
detector needs to be masked out due to spurious noise or damaged panels.
Sometimes, data from some detector panels must be completely ignored to reduce
computation time and thus allow rapid data collection.  If speckle pattern
classification can be accurately performed on only a fraction of a detector area,
it opens the door to solving the ``data reduction" problem that bottlenecks
high-throughput single particle imaging experiments.  That is to say, it can
save a considerable amount of time by eliminating the need for assembling and
calibrating all detector panels for the classification process.

\begin{figure}[htbp]
\includegraphics[width=0.5\textwidth,keepaspectratio]
{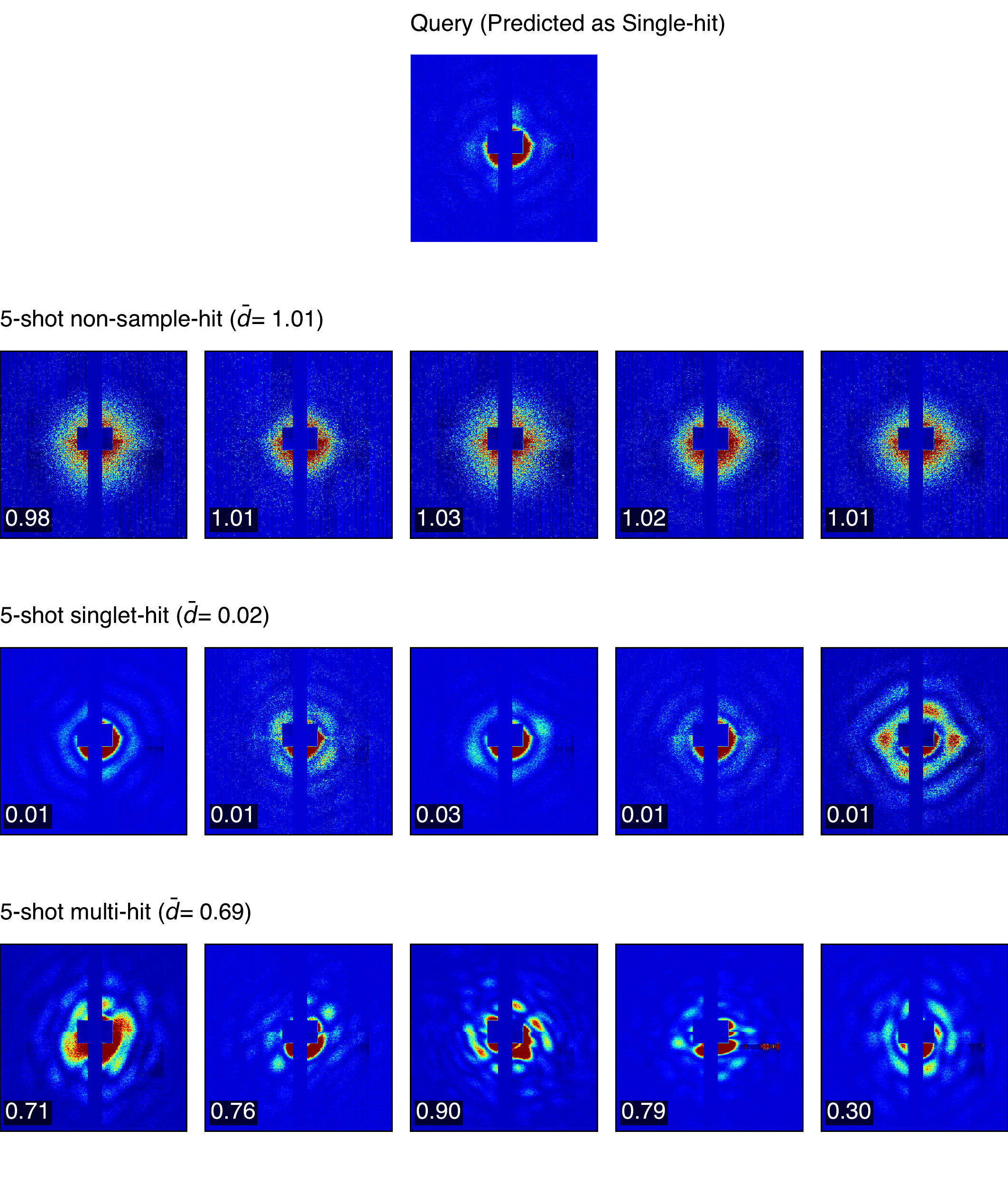}
\caption{3-way 5-shot classification on speckle patterns with 100\% detector
area available.}
\label{fig : 5 shot real single 100}
\end{figure}

Altogether, we randomly selected 345 single-hit and 655 non-single-hit speckle
patterns to form the test set, with non-single-hit made up of 331 non-sample-hit
and 324 multi-hit.  \SpeckleNN{} classifies speckle patterns in a 5-shot manner,
whereas Shi19 model uses a probability threshold of 0.9 for the classification
task.  The model accuracy and F-1 scores are summarized in Table \ref{tb : perf
100 vs 25}.  It is worth noting that data augmentation enhanced the accuracy of
Shi19 model significantly, from 83.8\% to 98\% when 100\% detector area is
available.  Meanwhile, under the same circumstances, \SpeckleNN{} and Shi19
model have the same accuracy and F-1 scores, respectively.  But \SpeckleNN{}
outperforms the competing model by a large margin when only 25\% detector area
is available.  Fig. \ref{fig : 5 shot real single 100} and Fig. \ref{fig : 5
shot real single 25} are demonstrations of 3-way 5-shot classification with
\SpeckleNN{} on speckle patterns with 100\% and 25\% detector area available,
respectively.  This result suggests that \SpeckleNN{} is a more robust speckle
pattern classifier and thus better suited for high-throughput single particle
imaging experiments.

\begin{table}[htbp]
\caption{Model accuracy (ACC) and F-1 scores in two scenarios: (1) 100\%
detector area is available; (2) 25\% detector area is available.  The percent
detector area visibility is annotated as a subscript in the table.  In
addition, we performed 5-shot classification using \SpeckleNN{}, and the
probability threshold used in
Shi19 model is 0.9.}
\label{tb : perf 100 vs 25}
\begin{tabular}{ l | l l | l l }
    Model        & ACC$_{100\%}$  &  F-1$_{100\%}$ & ACC$_{25\%}$  &  F-1$_{25\%}$  \\
    \SpeckleNN{} & 0.98           &  0.97          & 0.94          & 0.92      \\
    Shi19 model  & 0.98           &  0.97          & 0.74          & 0.64      \\
\end{tabular}
\end{table}

\begin{figure}[htbp]
\includegraphics[width=0.5\textwidth,keepaspectratio]
{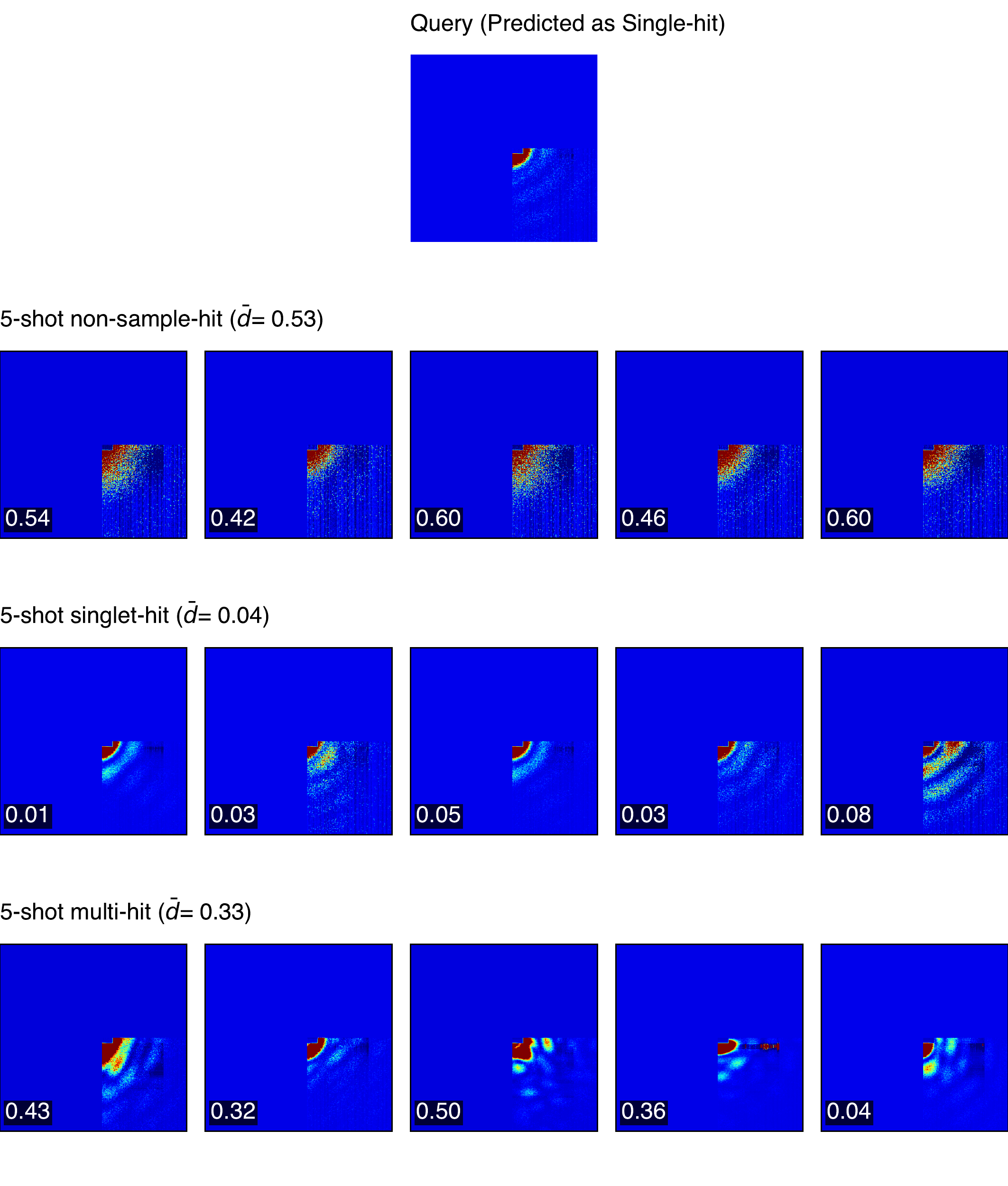}
\caption{3-way 5-shot classification on speckle patterns with 25\% detector area
available.}
\label{fig : 5 shot real single 25}
\end{figure}


\section{Conclusions}

In this work, we have introduced \SpeckleNN{}, a unified embedding model for
real-time speckle pattern classification in X-ray single particle imaging with
limited labeled examples.  The embedding model, trained with twin neural
networks, can directly map speckle patterns to a unified vector space, where
similarity is characterized by Euclidean distance.  We have provided two
distinct speckle pattern classification solutions.  Firstly, the model trained
on multiple samples offline allows few-shot classification of new never-seen
single-particle samples.  While our results show promising progress, future work
is needed to transfer the model trained on simulated data to real experimental
data.  A more realistic simulator \citep{ledigPhotoRealisticSingleImage2017} can
potentially bridge this gap.  Secondly, the model trained on one sample of
interest online exhibits notably improved performance in the presence of
substantial missing detector area, as compared to Shi19 model, a simple yet
effective neural network based single-particle classifier.  Our model's ability
to classify speckle patterns with partial detector information presents a
significant opportunity for the development of a rapid speckle pattern vetoing
process.  Additionally, data augmentation is crucial in both offline and online
training of our model, with a greater impact in online training.


\section*{Acknowledgment}

The project was conceived by J.B.T and C.H.Y. Earlier versions of the twin
neural network models were tested by E.F under the supervision of C.H.Y.  The
experiments were designed by C.W with inputs from C.H.Y.  C.W prepared the
simulation data, labeled experimental data, and trained and evaluated the
refined neural network model.  The manuscript was written by C.W and C.H.Y with
input from all authors.  This material is based upon work supported by the U.S.
Department of Energy, Office of Science, Office of Basic Energy Sciences under
Award Number FWP-100643.  Use of the Linac Coherent Light Source (LCLS), SLAC
National Accelerator Laboratory, is supported by the U.S. Department of Energy,
Office of Science, Office of Basic Energy Sciences under Contract No.
DE-AC02-76SF00515.

\bibliographystyle{plainnat}
\bibliography{bibliography.bib}

\begin{thebibliography}{34}
\providecommand{\natexlab}[1]{#1}
\providecommand{\url}[1]{\texttt{#1}}
\expandafter\ifx\csname urlstyle\endcsname\relax
  \providecommand{\doi}[1]{doi: #1}\else
  \providecommand{\doi}{doi: \begingroup \urlstyle{rm}\Url}\fi

\bibitem[Andreasson et~al.(2014)Andreasson, Martin, Liang, Timneanu, Aquila,
  Wang, Iwan, Svenda, Ekeberg, Hantke, Bielecki, Rolles, Rudenko, Foucar,
  Hartmann, Erk, Rudek, Chapman, Hajdu, and
  Barty]{andreassonAutomatedIdentificationClassification2014}
Jakob Andreasson, Andrew~V. Martin, Meng Liang, Nicusor Timneanu, Andrew
  Aquila, Fenglin Wang, Bianca Iwan, Martin Svenda, Tomas Ekeberg, Max Hantke,
  Johan Bielecki, Daniel Rolles, Artem Rudenko, Lutz Foucar, Robert Hartmann,
  Benjamin Erk, Benedikt Rudek, Henry~N. Chapman, Janos Hajdu, and Anton Barty.
\newblock Automated identification and classification of single particle serial
  femtosecond {{X-ray}} diffraction data.
\newblock \emph{Optics Express}, 22\penalty0 (3):\penalty0 2497, February 2014.
\newblock ISSN 1094-4087.
\newblock \doi{10.1364/OE.22.002497}.

\bibitem[Aquila et~al.(2015)Aquila, Barty, Bostedt, Boutet, Carini, {dePonte},
  Drell, Doniach, Downing, Earnest, Elmlund, Elser, G{\"u}hr, Hajdu, Hastings,
  {Hau-Riege}, Huang, Lattman, Maia, Marchesini, Ourmazd, Pellegrini, Santra,
  Schlichting, Schroer, Spence, Vartanyants, Wakatsuki, Weis, and
  Williams]{aquilaLinacCoherentLight2015}
A.~Aquila, A.~Barty, C.~Bostedt, S.~Boutet, G.~Carini, D.~{dePonte}, P.~Drell,
  S.~Doniach, K.~H. Downing, T.~Earnest, H.~Elmlund, V.~Elser, M.~G{\"u}hr,
  J.~Hajdu, J.~Hastings, S.~P. {Hau-Riege}, Z.~Huang, E.~E. Lattman, F.~R.
  N.~C. Maia, S.~Marchesini, A.~Ourmazd, C.~Pellegrini, R.~Santra,
  I.~Schlichting, C.~Schroer, J.~C.~H. Spence, I.~A. Vartanyants, S.~Wakatsuki,
  W.~I. Weis, and G.~J. Williams.
\newblock The linac coherent light source single particle imaging road map.
\newblock \emph{Structural Dynamics}, 2\penalty0 (4):\penalty0 041701, July
  2015.
\newblock ISSN 2329-7778.
\newblock \doi{10.1063/1.4918726}.

\bibitem[Ayyer et~al.(2016)Ayyer, Lan, Elser, and
  Loh]{ayyerDragonflyImplementationExpand2016}
Kartik Ayyer, Ti-Yen Lan, Veit Elser, and N.~Duane Loh.
\newblock {\emph{Dragonfly}} : An implementation of the expand\textendash
  maximize\textendash compress algorithm for single-particle imaging.
\newblock \emph{Journal of Applied Crystallography}, 49\penalty0 (4):\penalty0
  1320--1335, August 2016.
\newblock ISSN 1600-5767.
\newblock \doi{10.1107/S1600576716008165}.

\bibitem[Balestriero et~al.(2022)Balestriero, Misra, and
  LeCun]{balestrieroDataAugmentationWorthThousand2022}
Randall Balestriero, Ishan Misra, and Yann LeCun.
\newblock A {{Data-Augmentation Is Worth A Thousand Samples}}: {{Exact
  Quantification From Analytical Augmented Sample Moments}}, February 2022.

\bibitem[Bobkov et~al.(2015)Bobkov, Teslyuk, Kurta, Gorobtsov, Yefanov, Ilyin,
  Senin, and Vartanyants]{bobkovSortingAlgorithmsSingleparticle2015}
S.~A. Bobkov, A.~B. Teslyuk, R.~P. Kurta, O.~Yu. Gorobtsov, O.~M. Yefanov,
  V.~A. Ilyin, R.~A. Senin, and I.~A. Vartanyants.
\newblock Sorting algorithms for single-particle imaging experiments at
  {{X-ray}} free-electron lasers.
\newblock \emph{Journal of Synchrotron Radiation}, 22\penalty0 (6):\penalty0
  1345--1352, November 2015.
\newblock ISSN 1600-5775.
\newblock \doi{10.1107/S1600577515017348}.

\bibitem[Bromley et~al.(1993)Bromley, Bentz, Bottou, Guyon, Lecun, Moore,
  S{\"a}ckinger, and Shah]{bromleySignatureVerificationUsing1993}
Jane Bromley, James~W. Bentz, L{\'e}on Bottou, Isabelle Guyon, Yann Lecun,
  Cliff Moore, Eduard S{\"a}ckinger, and Roopak Shah.
\newblock {{SIGNATURE VERIFICATION USING A}} ``{{SIAMESE}}'' {{TIME DELAY
  NEURAL NETWORK}}.
\newblock \emph{International Journal of Pattern Recognition and Artificial
  Intelligence}, 07\penalty0 (04):\penalty0 669--688, August 1993.
\newblock ISSN 0218-0014, 1793-6381.
\newblock \doi{10.1142/S0218001493000339}.

\bibitem[Chang et~al.(2021)Chang, Slaughter, Mirchandaney, Donatelli, and
  Yoon]{changScalingAccelerationThreedimensional2021}
Hsing-Yin Chang, Elliott Slaughter, Seema Mirchandaney, Jeffrey Donatelli, and
  Chun~Hong Yoon.
\newblock Scaling and {{Acceleration}} of {{Three-dimensional Structure
  Determination}} for {{Single-Particle Imaging Experiments}} with
  {{SpiniFEL}}, September 2021.

\bibitem[Chapman et~al.(2006)Chapman, Barty, Bogan, Boutet, Frank, {Hau-Riege},
  Marchesini, Woods, Bajt, Benner, London, Pl{\"o}njes, Kuhlmann, Treusch,
  D{\"u}sterer, Tschentscher, Schneider, Spiller, M{\"o}ller, Bostedt, Hoener,
  Shapiro, Hodgson, {van der Spoel}, Burmeister, Bergh, Caleman, Huldt,
  Seibert, Maia, Lee, Sz{\"o}ke, Timneanu, and
  Hajdu]{chapmanFemtosecondDiffractiveImaging2006}
Henry~N. Chapman, Anton Barty, Michael~J. Bogan, S{\'e}bastien Boutet, Matthias
  Frank, Stefan~P. {Hau-Riege}, Stefano Marchesini, Bruce~W. Woods, Sa{\v s}a
  Bajt, W.~Henry Benner, Richard~A. London, Elke Pl{\"o}njes, Marion Kuhlmann,
  Rolf Treusch, Stefan D{\"u}sterer, Thomas Tschentscher, Jochen~R. Schneider,
  Eberhard Spiller, Thomas M{\"o}ller, Christoph Bostedt, Matthias Hoener,
  David~A. Shapiro, Keith~O. Hodgson, David {van der Spoel}, Florian
  Burmeister, Magnus Bergh, Carl Caleman, G{\"o}sta Huldt, M.~Marvin Seibert,
  Filipe R. N.~C. Maia, Richard~W. Lee, Abraham Sz{\"o}ke, Nicusor Timneanu,
  and Janos Hajdu.
\newblock Femtosecond diffractive imaging with a soft-{{X-ray}} free-electron
  laser.
\newblock \emph{Nature Physics}, 2\penalty0 (12):\penalty0 839--843, December
  2006.
\newblock ISSN 1745-2473, 1745-2481.
\newblock \doi{10.1038/nphys461}.

\bibitem[Chopra et~al.(2005)Chopra, Hadsell, and
  LeCun]{chopraLearningSimilarityMetric2005}
S.~Chopra, R.~Hadsell, and Y.~LeCun.
\newblock Learning a {{Similarity Metric Discriminatively}}, with
  {{Application}} to {{Face Verification}}.
\newblock In \emph{2005 {{IEEE Computer Society Conference}} on {{Computer
  Vision}} and {{Pattern Recognition}} ({{CVPR}}'05)}, volume~1, pages
  539--546, {San Diego, CA, USA}, 2005. {IEEE}.
\newblock ISBN 978-0-7695-2372-9.
\newblock \doi{10.1109/CVPR.2005.202}.

\bibitem[{Cruz-Ch{\'u}} et~al.(2021){Cruz-Ch{\'u}}, Hosseinizadeh, Mashayekhi,
  Fung, Ourmazd, and Schwander]{cruz-chuSelectingXFELSingleparticle2021}
Eduardo~R. {Cruz-Ch{\'u}}, Ahmad Hosseinizadeh, Ghoncheh Mashayekhi, Russell
  Fung, Abbas Ourmazd, and Peter Schwander.
\newblock Selecting {{XFEL}} single-particle snapshots by geometric machine
  learning.
\newblock \emph{Structural Dynamics}, 8\penalty0 (1):\penalty0 014701, January
  2021.
\newblock ISSN 2329-7778.
\newblock \doi{10.1063/4.0000060}.

\bibitem[Donatelli et~al.(2017)Donatelli, Sethian, and
  Zwart]{donatelliReconstructionLimitedSingleparticle2017}
Jeffrey~J. Donatelli, James~A. Sethian, and Peter~H. Zwart.
\newblock Reconstruction from limited single-particle diffraction data via
  simultaneous determination of state, orientation, intensity, and phase.
\newblock \emph{Proceedings of the National Academy of Sciences}, 114\penalty0
  (28):\penalty0 7222--7227, July 2017.
\newblock ISSN 0027-8424, 1091-6490.
\newblock \doi{10.1073/pnas.1708217114}.

\bibitem[Giannakis et~al.(2012)Giannakis, Schwander, and
  Ourmazd]{giannakisSymmetriesImageFormation2012}
Dimitrios Giannakis, Peter Schwander, and Abbas Ourmazd.
\newblock The symmetries of image formation by scattering {{I Theoretical}}
  framework.
\newblock \emph{Optics Express}, 20\penalty0 (12):\penalty0 12799, June 2012.
\newblock ISSN 1094-4087.
\newblock \doi{10.1364/OE.20.012799}.

\bibitem[Hajdu et~al.(2016)Hajdu, Williams, Song, Mancuso, Boutet, Elser,
  Aquila, Nakagawa, Fromme, Kirian, Loh, Munke, Westphal, Sierra, Hantke,
  Timneanu, Bielecki, M{\"u}hlig, Andreasson, Daurer, Kumar, Larsson, Maia,
  Nettelblad, Okamoto, Seibert, Svenda, Schot, Awel, Ayyer, Barty, Wiedorn,
  Xavier, Bean, Berntsen, DeMirci, Higashiura, Hogue, Kim, {Ti-Yen Lan}, Nam,
  Nelson, Ourmazd, Rose, Schwander, Vartanyants, Yoon, Zook, Bucher, Liu,
  Chapman, Hosseinzadeh, and Sellberg]{janoshajduCoherentDiffractionSingle2016}
Janos Hajdu, Garth~J. Williams, Changyong Song, Adrian Mancuso, S{\'e}bastien
  Boutet, Veit Elser, Andrew Aquila, Atsushi Nakagawa, Petra Fromme, Richard
  Kirian, Duane Loh, Anna Munke, Daniel Westphal, Raymond~G. Sierra, Max~F.
  Hantke, Nicusor Timneanu, Johan Bielecki, Kerstin M{\"u}hlig, Jakob
  Andreasson, Benedikt Daurer, Hemanth Kumar, Daniel S.~D. Larsson, Filipe R.
  N.~C Maia, Carl Nettelblad, Kenta Okamoto, Marvin Seibert, Martin Svenda,
  Gijs Van~Der Schot, Salah Awel, Kartik Ayyer, Anton Barty, Max~O. Wiedorn,
  Paulraj~Lourdu Xavier, Richard~J. Bean, Peter Berntsen, Hasan DeMirci,
  Akifumi Higashiura, Brenda Hogue, Yoonhee Kim, {Ti-Yen Lan}, Daewoong Nam,
  Garrett Nelson, Abbas Ourmazd, Max Rose, Peter Schwander, Ivan~A.
  Vartanyants, Chun~Hong Yoon, James Zook, Maximilian Bucher, Haiguang Liu,
  Henry Chapman, Ahmad Hosseinzadeh, and Jonas~A. Sellberg.
\newblock Coherent diffraction of single {{Rice Dwarf Virus}} particles using
  hard {{X-rays}} at the {{Linac Coherent Light Source}}.
\newblock 2016.
\newblock \doi{10.6084/M9.FIGSHARE.C.2342581}.

\bibitem[Hoffer and Ailon(2014)]{hofferDeepMetricLearning2014}
Elad Hoffer and Nir Ailon.
\newblock Deep metric learning using {{Triplet}} network.
\newblock 2014.
\newblock \doi{10.48550/ARXIV.1412.6622}.

\bibitem[Ignatenko et~al.(2021)Ignatenko, Assalauova, Bobkov, Gelisio, Teslyuk,
  Ilyin, and Vartanyants]{ignatenkoClassificationDiffractionPatterns2021}
Alexandr Ignatenko, Dameli Assalauova, Sergey~A Bobkov, Luca Gelisio, Anton~B
  Teslyuk, Viacheslav~A Ilyin, and Ivan~A Vartanyants.
\newblock Classification of diffraction patterns in single particle imaging
  experiments performed at x-ray free-electron lasers using a convolutional
  neural network.
\newblock \emph{Machine Learning: Science and Technology}, 2\penalty0
  (2):\penalty0 025014, June 2021.
\newblock ISSN 2632-2153.
\newblock \doi{10.1088/2632-2153/abd916}.

\bibitem[Kapoor and Narayanan(2022)]{kapoorLeakageReproducibilityCrisis2022}
Sayash Kapoor and Arvind Narayanan.
\newblock Leakage and the {{Reproducibility Crisis}} in {{ML-based Science}}.
\newblock \emph{arXiv}, 2022.
\newblock \doi{10.48550/ARXIV.2207.07048}.

\bibitem[Kingma and Ba(2017)]{kingmaAdamMethodStochastic2017}
Diederik~P. Kingma and Jimmy Ba.
\newblock Adam: {{A Method}} for {{Stochastic Optimization}}.
\newblock \emph{arXiv:1412.6980 [cs]}, January 2017.

\bibitem[Ledig et~al.(2017)Ledig, Theis, Huszar, Caballero, Cunningham, Acosta,
  Aitken, Tejani, Totz, Wang, and Shi]{ledigPhotoRealisticSingleImage2017}
Christian Ledig, Lucas Theis, Ferenc Huszar, Jose Caballero, Andrew Cunningham,
  Alejandro Acosta, Andrew Aitken, Alykhan Tejani, Johannes Totz, Zehan Wang,
  and Wenzhe Shi.
\newblock Photo-{{Realistic Single Image Super-Resolution Using}} a
  {{Generative Adversarial Network}}, May 2017.

\bibitem[Li(2020)]{liDiffractionDataAerosolized2020}
Haoyuan Li.
\newblock Diffraction data from aerosolized {{Coliphage PR772}} virus particles
  imaged with the {{Linac Coherent Light Source}}, 2020.

\bibitem[Li et~al.(2022)Li, MacArthur, Littleton, Dunne, Huang, and
  Zhu]{liFemtosecondTerawattHardXRay2022}
Haoyuan Li, James MacArthur, Sean Littleton, Mike Dunne, Zhirong Huang, and
  Diling Zhu.
\newblock Femtosecond-{{Terawatt Hard X-Ray Pulse Generation}} with {{Chirped
  Pulse Amplification}} on a {{Free Electron Laser}}.
\newblock \emph{Physical Review Letters}, 129\penalty0 (21):\penalty0 213901,
  November 2022.
\newblock ISSN 0031-9007, 1079-7114.
\newblock \doi{10.1103/PhysRevLett.129.213901}.

\bibitem[Loh and Elser(2009)]{lohReconstructionAlgorithmSingleparticle2009}
Ne-Te~Duane Loh and Veit Elser.
\newblock Reconstruction algorithm for single-particle diffraction imaging
  experiments.
\newblock \emph{Physical Review E}, 80\penalty0 (2):\penalty0 026705, August
  2009.
\newblock ISSN 1539-3755, 1550-2376.
\newblock \doi{10.1103/PhysRevE.80.026705}.

\bibitem[Neutze et~al.(2000)Neutze, Wouts, {van der Spoel}, Weckert, and
  Hajdu]{neutzePotentialBiomolecularImaging2000}
Richard Neutze, Remco Wouts, David {van der Spoel}, Edgar Weckert, and Janos
  Hajdu.
\newblock Potential for biomolecular imaging with femtosecond {{X-ray}} pulses.
\newblock \emph{Nature}, 406\penalty0 (6797):\penalty0 752--757, August 2000.
\newblock ISSN 0028-0836, 1476-4687.
\newblock \doi{10.1038/35021099}.

\bibitem[Peck et~al.(2022)Peck, Chang, Dujardin, Ramalingam, Uervirojnangkoorn,
  Wang, Mancuso, Poitevin, and Yoon]{peckSkopiSimulationPackage2022}
Ariana Peck, Hsing-Yin Chang, Antoine Dujardin, Deeban Ramalingam, Monarin
  Uervirojnangkoorn, Zhaoyou Wang, Adrian Mancuso, Fr{\'e}d{\'e}ric Poitevin,
  and Chun~Hong Yoon.
\newblock {\emph{Skopi}} : A simulation package for diffractive imaging of
  noncrystalline biomolecules.
\newblock \emph{Journal of Applied Crystallography}, 55\penalty0 (4), August
  2022.
\newblock ISSN 1600-5767.
\newblock \doi{10.1107/S1600576722005994}.

\bibitem[Reddy et~al.(2017)Reddy, Yoon, Aquila, Awel, Ayyer, Barty, Berntsen,
  Bielecki, Bobkov, Bucher, Carini, Carron, Chapman, Daurer, DeMirci, Ekeberg,
  Fromme, Hajdu, Hanke, Hart, Hogue, Hosseinizadeh, Kim, Kirian, Kurta,
  Larsson, Duane~Loh, Maia, Mancuso, M{\"u}hlig, Munke, Nam, Nettelblad,
  Ourmazd, Rose, Schwander, Seibert, Sellberg, Song, Spence, Svenda, {Van der
  Schot}, Vartanyants, Williams, and Xavier]{reddyCoherentSoftXray2017a}
Hemanth~K.N. Reddy, Chun~Hong Yoon, Andrew Aquila, Salah Awel, Kartik Ayyer,
  Anton Barty, Peter Berntsen, Johan Bielecki, Sergey Bobkov, Maximilian
  Bucher, Gabriella~A. Carini, Sebastian Carron, Henry Chapman, Benedikt
  Daurer, Hasan DeMirci, Tomas Ekeberg, Petra Fromme, Janos Hajdu, Max~Felix
  Hanke, Philip Hart, Brenda~G. Hogue, Ahmad Hosseinizadeh, Yoonhee Kim,
  Richard~A. Kirian, Ruslan~P. Kurta, Daniel~S.D. Larsson, N.~Duane~Loh,
  Filipe~R.N.C. Maia, Adrian~P. Mancuso, Kerstin M{\"u}hlig, Anna Munke,
  Daewoong Nam, Carl Nettelblad, Abbas Ourmazd, Max Rose, Peter Schwander,
  Marvin Seibert, Jonas~A. Sellberg, Changyong Song, John~C.H. Spence, Martin
  Svenda, Gijs {Van der Schot}, Ivan~A. Vartanyants, Garth~J. Williams, and
  P.~Lourdu Xavier.
\newblock Coherent soft {{X-ray}} diffraction imaging of coliphage {{PR772}} at
  the {{Linac}} coherent light source.
\newblock \emph{Scientific Data}, 4\penalty0 (1):\penalty0 170079, December
  2017.
\newblock ISSN 2052-4463.
\newblock \doi{10.1038/sdata.2017.79}.

\bibitem[Redmon and Farhadi(2016)]{redmonYOLO9000BetterFaster2016}
Joseph Redmon and Ali Farhadi.
\newblock {{YOLO9000}}: {{Better}}, {{Faster}}, {{Stronger}}.
\newblock \emph{arXiv:1612.08242 [cs]}, December 2016.

\bibitem[Redmon and Farhadi(2018)]{redmonYOLOv3IncrementalImprovement2018}
Joseph Redmon and Ali Farhadi.
\newblock {{YOLOv3}}: {{An Incremental Improvement}}.
\newblock \emph{arXiv}, page~6, 2018.

\bibitem[Schroff et~al.(2015)Schroff, Kalenichenko, and
  Philbin]{schroffFaceNetUnifiedEmbedding2015}
Florian Schroff, Dmitry Kalenichenko, and James Philbin.
\newblock {{FaceNet}}: {{A}} unified embedding for face recognition and
  clustering.
\newblock In \emph{2015 {{IEEE Conference}} on {{Computer Vision}} and
  {{Pattern Recognition}} ({{CVPR}})}, pages 815--823, {Boston, MA, USA}, June
  2015. {IEEE}.
\newblock ISBN 978-1-4673-6964-0.
\newblock \doi{10.1109/CVPR.2015.7298682}.

\bibitem[Schwander et~al.(2012)Schwander, Giannakis, Yoon, and
  Ourmazd]{schwanderSymmetriesImageFormation2012}
Peter Schwander, Dimitrios Giannakis, Chun~Hong Yoon, and Abbas Ourmazd.
\newblock The symmetries of image formation by scattering {{II Applications}}.
\newblock \emph{Optics Express}, 20\penalty0 (12):\penalty0 12827, June 2012.
\newblock ISSN 1094-4087.
\newblock \doi{10.1364/OE.20.012827}.

\bibitem[Seibert et~al.(2011)Seibert, Ekeberg, Maia, Svenda, Andreasson,
  J{\"o}nsson, Odi{\'c}, Iwan, Rocker, Westphal, Hantke, DePonte, Barty,
  Schulz, Gumprecht, Coppola, Aquila, Liang, White, Martin, Caleman, Stern,
  Abergel, Seltzer, Claverie, Bostedt, Bozek, Boutet, Miahnahri, Messerschmidt,
  Krzywinski, Williams, Hodgson, Bogan, Hampton, Sierra, Starodub, Andersson,
  Bajt, Barthelmess, Spence, Fromme, Weierstall, Kirian, Hunter, Doak,
  Marchesini, {Hau-Riege}, Frank, Shoeman, Lomb, Epp, Hartmann, Rolles,
  Rudenko, Schmidt, Foucar, Kimmel, Holl, Rudek, Erk, H{\"o}mke, Reich,
  Pietschner, Weidenspointner, Str{\"u}der, Hauser, Gorke, Ullrich,
  Schlichting, Herrmann, Schaller, Schopper, Soltau, K{\"u}hnel, Andritschke,
  Schr{\"o}ter, Krasniqi, Bott, Schorb, Rupp, Adolph, Gorkhover, Hirsemann,
  Potdevin, Graafsma, Nilsson, Chapman, and
  Hajdu]{seibertSingleMimivirusParticles2011}
M.~Marvin Seibert, Tomas Ekeberg, Filipe R. N.~C. Maia, Martin Svenda, Jakob
  Andreasson, Olof J{\"o}nsson, Du{\v s}ko Odi{\'c}, Bianca Iwan, Andrea
  Rocker, Daniel Westphal, Max Hantke, Daniel~P. DePonte, Anton Barty, Joachim
  Schulz, Lars Gumprecht, Nicola Coppola, Andrew Aquila, Mengning Liang,
  Thomas~A. White, Andrew Martin, Carl Caleman, Stephan Stern, Chantal Abergel,
  Virginie Seltzer, Jean-Michel Claverie, Christoph Bostedt, John~D. Bozek,
  S{\'e}bastien Boutet, A.~Alan Miahnahri, Marc Messerschmidt, Jacek
  Krzywinski, Garth Williams, Keith~O. Hodgson, Michael~J. Bogan, Christina~Y.
  Hampton, Raymond~G. Sierra, Dmitri Starodub, Inger Andersson, Sa{\v s}a Bajt,
  Miriam Barthelmess, John C.~H. Spence, Petra Fromme, Uwe Weierstall, Richard
  Kirian, Mark Hunter, R.~Bruce Doak, Stefano Marchesini, Stefan~P.
  {Hau-Riege}, Matthias Frank, Robert~L. Shoeman, Lukas Lomb, Sascha~W. Epp,
  Robert Hartmann, Daniel Rolles, Artem Rudenko, Carlo Schmidt, Lutz Foucar,
  Nils Kimmel, Peter Holl, Benedikt Rudek, Benjamin Erk, Andr{\'e} H{\"o}mke,
  Christian Reich, Daniel Pietschner, Georg Weidenspointner, Lothar
  Str{\"u}der, G{\"u}nter Hauser, Hubert Gorke, Joachim Ullrich, Ilme
  Schlichting, Sven Herrmann, Gerhard Schaller, Florian Schopper, Heike Soltau,
  Kai-Uwe K{\"u}hnel, Robert Andritschke, Claus-Dieter Schr{\"o}ter, Faton
  Krasniqi, Mario Bott, Sebastian Schorb, Daniela Rupp, Marcus Adolph, Tais
  Gorkhover, Helmut Hirsemann, Guillaume Potdevin, Heinz Graafsma, Bj{\"o}rn
  Nilsson, Henry~N. Chapman, and Janos Hajdu.
\newblock Single mimivirus particles intercepted and imaged with an {{X-ray}}
  laser.
\newblock \emph{Nature}, 470\penalty0 (7332):\penalty0 78--81, February 2011.
\newblock ISSN 0028-0836, 1476-4687.
\newblock \doi{10.1038/nature09748}.

\bibitem[Shi et~al.(2019)Shi, Yin, Tai, DeMirci, Hosseinizadeh, Hogue, Li,
  Ourmazd, Schwander, Vartanyants, Yoon, Aquila, and
  Liu]{shiEvaluationPerformanceClassification2019}
Yingchen Shi, Ke~Yin, Xuecheng Tai, Hasan DeMirci, Ahmad Hosseinizadeh,
  Brenda~G. Hogue, Haoyuan Li, Abbas Ourmazd, Peter Schwander, Ivan~A.
  Vartanyants, Chun~Hong Yoon, Andrew Aquila, and Haiguang Liu.
\newblock Evaluation of the performance of classification algorithms for
  {{XFEL}} single-particle imaging data.
\newblock \emph{IUCrJ}, 6\penalty0 (2):\penalty0 331--340, March 2019.
\newblock ISSN 2052-2525.
\newblock \doi{10.1107/S2052252519001854}.

\bibitem[Snell et~al.(2017)Snell, Swersky, and
  Zemel]{snellPrototypicalNetworksFewshot2017}
Jake Snell, Kevin Swersky, and Richard~S. Zemel.
\newblock Prototypical {{Networks}} for {{Few-shot Learning}}.
\newblock \emph{arXiv:1703.05175 [cs, stat]}, June 2017.

\bibitem[Vinyals et~al.(2017)Vinyals, Blundell, Lillicrap, Kavukcuoglu, and
  Wierstra]{vinyalsMatchingNetworksOne2017}
Oriol Vinyals, Charles Blundell, Timothy Lillicrap, Koray Kavukcuoglu, and Daan
  Wierstra.
\newblock Matching {{Networks}} for {{One Shot Learning}}, December 2017.

\bibitem[Yoon et~al.(2011)Yoon, Schwander, Abergel, Andersson, Andreasson,
  Aquila, Bajt, Barthelmess, Barty, Bogan, Bostedt, Bozek, Chapman, Claverie,
  Coppola, DePonte, Ekeberg, Epp, Erk, Fleckenstein, Foucar, Graafsma,
  Gumprecht, Hajdu, Hampton, Hartmann, Hartmann, Hartmann, Hauser, Hirsemann,
  Holl, Kassemeyer, Kimmel, Kiskinova, Liang, Loh, Lomb, Maia, Martin, Nass,
  Pedersoli, Reich, Rolles, Rudek, Rudenko, Schlichting, Schulz, Seibert,
  Seltzer, Shoeman, Sierra, Soltau, Starodub, Steinbrener, Stier, Str{\"u}der,
  Svenda, Ullrich, Weidenspointner, White, Wunderer, and
  Ourmazd]{yoonUnsupervisedClassificationSingleparticle2011}
Chun~Hong Yoon, Peter Schwander, Chantal Abergel, Inger Andersson, Jakob
  Andreasson, Andrew Aquila, Sa{\v s}a Bajt, Miriam Barthelmess, Anton Barty,
  Michael~J. Bogan, Christoph Bostedt, John Bozek, Henry~N. Chapman,
  Jean-Michel Claverie, Nicola Coppola, Daniel~P. DePonte, Tomas Ekeberg,
  Sascha~W. Epp, Benjamin Erk, Holger Fleckenstein, Lutz Foucar, Heinz
  Graafsma, Lars Gumprecht, Janos Hajdu, Christina~Y. Hampton, Andreas
  Hartmann, Elisabeth Hartmann, Robert Hartmann, Gunter Hauser, Helmut
  Hirsemann, Peter Holl, Stephan Kassemeyer, Nils Kimmel, Maya Kiskinova,
  Mengning Liang, Ne-Te~Duane Loh, Lukas Lomb, Filipe R. N.~C. Maia, Andrew~V.
  Martin, Karol Nass, Emanuele Pedersoli, Christian Reich, Daniel Rolles,
  Benedikt Rudek, Artem Rudenko, Ilme Schlichting, Joachim Schulz, Marvin
  Seibert, Virginie Seltzer, Robert~L. Shoeman, Raymond~G. Sierra, Heike
  Soltau, Dmitri Starodub, Jan Steinbrener, Gunter Stier, Lothar Str{\"u}der,
  Martin Svenda, Joachim Ullrich, Georg Weidenspointner, Thomas~A. White,
  Cornelia Wunderer, and Abbas Ourmazd.
\newblock Unsupervised classification of single-particle {{X-ray}} diffraction
  snapshots by spectral clustering.
\newblock \emph{Optics Express}, 19\penalty0 (17):\penalty0 16542, August 2011.
\newblock ISSN 1094-4087.
\newblock \doi{10.1364/OE.19.016542}.

\bibitem[Yoon(2012)]{yoonNovelAlgorithmsCoherent2012}
Chunhong Yoon.
\newblock Novel algorithms in coherent diffraction imaging using x-ray
  free-electron lasers.
\newblock In Philip~J. Bones, Michael~A. Fiddy, and Rick~P. Millane, editors,
  \emph{{{SPIE Optical Engineering}} + {{Applications}}}, page 85000H, {San
  Diego, California, USA}, October 2012.
\newblock \doi{10.1117/12.953634}.

\end{thebibliography}

\end{document}